%% file: main.tex
\documentclass[10pt]{article}

\usepackage{arxiv}
\usepackage[utf8]{inputenc}
\usepackage[T1]{fontenc}
\usepackage{amsmath,amssymb}
\usepackage{graphicx}
\usepackage{booktabs}
\usepackage{array}
\usepackage{multirow}
\usepackage{enumitem}
\usepackage{siunitx}
\usepackage{pifont}
\usepackage{subcaption}
\usepackage{colortbl}
\usepackage{xcolor}
\usepackage[numbers]{natbib}
\usepackage{microtype}
\usepackage{placeins}
\usepackage{hyperref}
\usepackage{url}

\definecolor{macRose}{HTML}{F8A5B3}
\definecolor{macLav}{HTML}{C2B5EF}
\definecolor{macMint}{HTML}{80D4BE}
\definecolor{macPeach}{HTML}{FBBE8C}
\definecolor{propbg}{RGB}{241,232,214}
\definecolor{openbg}{RGB}{247,241,221}
\definecolor{threeDbg}{RGB}{221,231,245}
\definecolor{myorange}{HTML}{FDE4D0}
\definecolor{myyellow}{HTML}{FFF2CC}
\definecolor{myblue}{HTML}{D9E1F2}
\definecolor{mygreen}{HTML}{D6F0D6}
\definecolor{mygreens}{RGB}{34,139,34}
\definecolor{myred}{RGB}{220,20,60}
\definecolor{baselinegray}{HTML}{F2F2F2}

\newcommand{\cmark}{\textcolor{mygreens}{\ding{51}}}
\newcommand{\xmark}{\textcolor{myred}{\ding{55}}}

\newcommand{\uline}[1]{\underline{#1}}

\hypersetup{
  colorlinks=true,
  linkcolor=blue,
  citecolor=green!50!black,
  urlcolor=blue,
  pdftitle={PointQ-Bench: Benchmarking Diagnostic and Interpretable Point Cloud Quality Assessment},
  pdfauthor={Duanchu Wang, Cheng Li, Junjie Yang, Jing Huang, Zihang Cheng, Zhi Gao, Bohong Zhu, Di Wang},
  pdfkeywords={Point cloud quality assessment, Quality reasoning, MLLM, 3D vision-language models, No-reference assessment}
}

\title{PointQ-Bench: Benchmarking Diagnostic and Interpretable Point Cloud Quality Assessment}

\author{%
\textbf{Duanchu Wang$^{1,*}$}\quad
\textbf{Cheng Li$^{1,*}$}\quad
\textbf{Junjie Yang$^{2}$}\quad
\textbf{Jing Huang$^{1}$}\quad
\textbf{Zihang Cheng$^{1}$}\\
Zhi Gao$^{1}$\quad Bohong Zhu$^{1}$\quad Di Wang$^{3,4,\dagger}$\\[0.7em]
$^{1}$Xi'an Jiaotong University, Xi'an, Shaanxi, China\\
$^{2}$University of Chinese Academy of Sciences, Shenyang, China\\
$^{3}$School of Software Engineering, Xi'an Jiaotong University, Xi'an, Shaanxi, China\\
$^{4}$State Key Laboratory of Human-Machine Hybrid Augmented Intelligence, Xi'an Jiaotong University\\[0.7em]
$^{*}$Equal contribution\qquad $^{\dagger}$Corresponding author\\
\texttt{diwang@xjtu.edu.cn}
}
\date{}

\renewcommand{\headeright}{arXiv preprint}

\renewcommand{\shorttitle}{PointQ-Bench}

\begin{document}
\maketitle

\begin{abstract}
Point cloud quality plays a critical role in 3D acquisition, reconstruction, rendering, and perception, yet existing point cloud quality assessment (PCQA) research remains largely centered on scalar score prediction. In practical inspection scenarios, quality assessment often involves identifying defects, characterizing dominant issue types, assessing downstream usability, and providing evidence-supported descriptions, which are not explicitly evaluated by current benchmarks. We introduce PointQ-Bench, a benchmark designed to extend PCQA from scalar scoring toward comprehensive quality understanding. PointQ-Bench consists of 3,083 point clouds spanning authentic scans, simulated distortions, and AI-generated content, covering eight major issue types. Each sample is annotated with mean opinion scores (MOS), quality levels, issue tags, expert-grounded descriptions, and 12,332 question–answer pairs. The benchmark supports three perception-oriented tasks—anomaly sensing, defect diagnosis, and usability grading, as well as a cognition-oriented task of open-ended quality reporting. To evaluate free-form quality descriptions, we further propose SSFRQ-5D, a five-dimensional evaluation protocol validated through human–AI agreement analysis. Extensive experiments on 14 vision–language models and traditional PCQA baselines reveal a consistent perception–diagnosis gap: while current models exhibit emerging abilities in coarse defect perception, they struggle with grounded diagnosis and quality calibration. Strong 2D MLLMs generally outperform existing 3D VLMs, and the benefit of additional views or point-level inputs is non-uniform, varying across tasks, data sources, and models—particularly under boundary-ambiguous conditions. Overall, PointQ-Bench provides a diagnostic testbed for advancing reliable and interpretable point cloud quality understanding.
Project repository: \url{https://github.com/Mujeco/PointQ-Bench}.
\end{abstract}

\keywords{Point cloud quality assessment $\cdot$ Quality reasoning $\cdot$ MLLM $\cdot$ 3D vision-language models $\cdot$ No-reference assessment}
\clearpage

\begin{figure}[t]
  \centering
  \includegraphics[width=\textwidth]{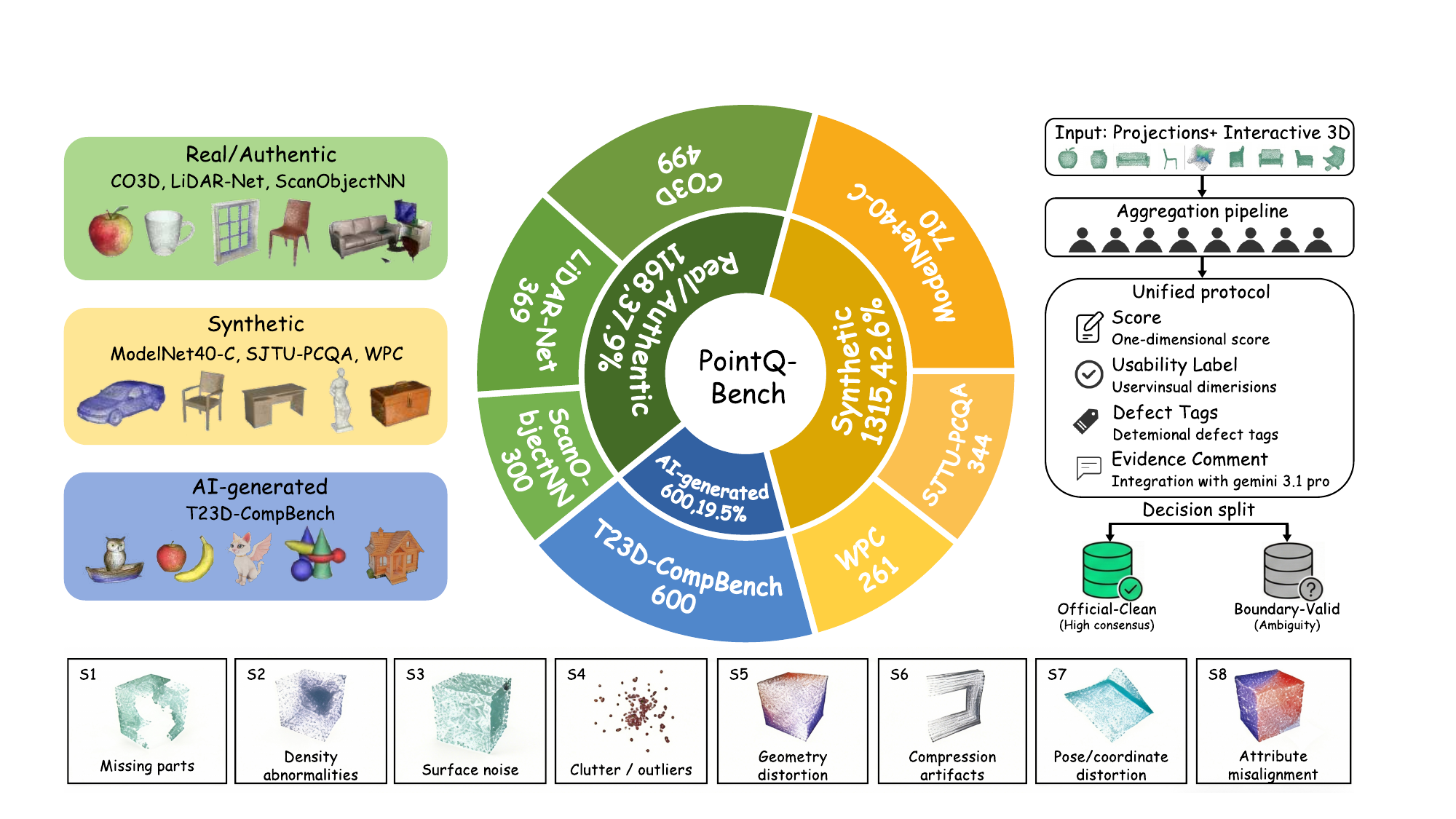}

  \caption{Overview of PointQ-Bench. The benchmark integrates
  authentic, simulated, and AI-generated data under a unified
  annotation protocol, extending scalar MOS assessment with an
  explicit defect taxonomy and ambiguity-aware decision splits.}
\label{fig:teaser}
\end{figure}

\begin{table}[t] 

\centering
\captionsetup{skip=1pt}
\caption{Comparison of PointQ-Bench with existing point cloud quality datasets.}
\vspace{2mm}
\label{tab:dataset_comparison}

\setlength{\tabcolsep}{5pt}

\renewcommand{\arraystretch}{1.10}

\resizebox{\textwidth}{!}{

\begin{tabular}{l c c c c c c c c c c l}

\toprule

\multirow{2}{*}{Dataset} & \multirow{2}{*}{Task} & \multirow{2}{*}{Size}

& \multicolumn{3}{c}{Source Types}

& \multicolumn{5}{c}{Annotation Fields}

& \multirow{2}{*}{Remarks} \\

\cmidrule(lr){4-6}\cmidrule(lr){7-11}

& & & Real & Syn. & AI & Full Hum. & MOS & Qual. Lvls & Issue Tags & Quality Desc. & \\

\midrule

SJTU-PCQA~\cite{sjtupcqa2020}   & PCQA    & 420    & \xmark & \cmark & \xmark & \cmark & \cmark & \xmark & \xmark & \xmark & Synthetic distortions only \\

WPC~\cite{wpc2023}              & PCQA    & 740    & \xmark & \cmark & \xmark & \cmark & \cmark & \xmark & \xmark & \xmark & Compression / noise / downsampling \\

SIAT-PCQD~\cite{wu2021siatpcqd} & VR-PCQA & 340    & \xmark & \cmark & \xmark & \cmark & \cmark & \xmark & \xmark & \xmark & 6DoF HMD-based setting \\

BASICS~\cite{ak2023basics}      & PCQA    & $\sim$1500 & \xmark & \cmark & \xmark & \cmark & \cmark & \xmark & \xmark & \xmark & Large-scale compression study \\

LS-PCQA~\cite{lspcqa2020}       & NR-PCQA & 22568  & \xmark & \cmark & \xmark & \xmark & \xmark & \xmark & \xmark & \xmark & Large-scale release is not fully human-annotated \\

\rowcolor{green!8}

\textbf{PointQ-Bench (Ours)} & \textbf{Quality Understanding} & \textbf{3083}

& \textbf{\cmark} & \textbf{\cmark} & \textbf{\cmark}

& \textbf{\cmark} & \textbf{\cmark} & \textbf{\cmark} & \textbf{\cmark} & \textbf{\cmark}

& \textbf{8 raters/sample; authentic + simulated + AI-generated} \\

\bottomrule

\end{tabular}

}

\end{table}

\section{Introduction}

Point clouds have become a fundamental representation for 3D content in acquisition, reconstruction, rendering, and perception pipelines, and are widely used in applications such as autonomous driving, robotics, digital twins, and immersive media~\cite{guo2021dlpointcloudsurvey, wu2021siatpcqd}. In real-world scenarios, however, point clouds are inevitably degraded by
sensing noise, reconstruction errors, compression artifacts, and transmission
loss. Modern generative pipelines introduce additional failure
modes~\cite{he2023t3bench,zhang2024mate3d,t23dcompbench2025}. Such degradations not only reduce perceptual fidelity, but also undermine the reliability of downstream 3D understanding and decision-making. As a result, reliable point cloud quality assessment (PCQA) is increasingly important for both practical 3D systems and multimodal perception research.

Existing PCQA methods can be broadly categorized into full-reference~\cite{meynet2020pcqm,zhang2021ms,yang2022mped}, reduced-reference~\cite{zhou2023rrcap}, and no-reference approaches~\cite{yang2022itpcqa,shan2023gpanet}. Because reference point clouds are often unavailable in practice, no-reference PCQA is the most realistic setting. Over the years, no-reference methods have evolved from point-only predictors to multimodal fusion models, and more recently to language-assisted or vision-language predictors ~\cite{zhou2023pcqasurvey}. Despite these architectural advances, however, the dominant paradigm remains scalar quality prediction—typically formulated as  mean opinion score (MOS) regression or opinion-score distribution prediction. While such formulations are effective for correlation-based evaluation, they fall short of practical inspection requirements.  In real-world deployment, a quality assessment system should not only output a score, but also detect noticeable defects, identify major issue types, judge downstream usability, and explain the supporting evidence. The same limitation appears in current benchmarks: public datasets such as SJTU-PCQA~\cite{sjtupcqa2020}, SIAT-PCQD~\cite{wu2021siatpcqd}, and WPC~\cite{wpc2023} have advanced PCQA research, yet they primarily evaluate MOS estimation. As summarized in Table~\ref{tab:dataset_comparison}, existing benchmarks offer limited support for diagnosing quality issues or assessing decision-oriented and explainable quality reasoning.

To address this critical gap, we present \textbf{PointQ-Bench}, to the best of our knowledge, the first systematic benchmark for point-cloud quality understanding beyond scalar prediction. PointQ-Bench contains 3,083 point clouds from three complementary source families—authentic scans, simulated distortions, and AI-generated content—covering eight major issue types. In addition to MOS, it provides quality levels, multi-label issue tags, expert-grounded descriptions, and 12,332 question-answer pairs. The benchmark supports three perception tasks (anomaly sensing, defect diagnosis, and usability grading) together with one reasoning task of open-ended quality reporting. To make such open-ended reports reliably comparable, we further introduce SSFRQ-5D, a lightweight five-dimensional evaluation protocol validated through human--AI agreement analysis.

Extensive experiments on 14 state-of-the-art vision-language models and traditional PCQA baselines reveal that current models already exhibit emerging quality-understanding abilities. However, their main limitation lies in grounded diagnosis and quality calibration rather than coarse defect perception. Strong 2D MLLMs generally outperform current 3D VLMs, while the benefit of additional views or points is highly non-uniform and depends on task, source family, and model—especially under boundary-ambiguous conditions. These findings suggest that the field should move beyond scalar prediction toward a broader and more reliable paradigm of point-cloud quality understanding, especially for grounded diagnosis and decision calibration under boundary ambiguity.

Our main contributions are summarized as follows:
\begin{itemize}[leftmargin=*]
    \item We introduce PointQ-Bench, the first benchmark explicitly designed for point-cloud quality understanding beyond scalar prediction. It provides a comprehensive evaluation platform containing 3,083 point clouds and 12,332 question-answer pairs across eight major issue types and three complementary source families.
    \item We establish a unified multi-task framework that goes beyond conventional MOS-only evaluation, covering problem existence judgment, defect typing, usability grading, and standardized open-ended quality reasoning.
    \item We propose SSFRQ-5D, a lightweight multidimensional judging protocol for open-ended quality descriptions, and validate its effectiveness through human--machine agreement analysis against domain experts.
    \item We systematically benchmark 14 state-of-the-art proprietary MLLMs, open-source MLLMs, native 3D VLMs, and traditional PCQA baselines. The results identify grounded diagnosis and quality calibration as major bottlenecks, while the value of additional views or points remains non-uniform across tasks, source families, and models.
    
\end{itemize}

\section{Related Work}

\subsection{Point Cloud Quality Assessment}

PCQA has traditionally been formulated as a perceptual quality prediction problem for distorted point clouds. Early research primarily focused on full-reference quality metrics, where perceptual fidelity is measured by comparing a distorted point cloud against its pristine reference. Representative methods such as MS-GraphSIM~\cite{zhang2021ms}, PCQM~\cite{meynet2020pcqm}, and MPED~\cite{yang2022mped} demonstrate that perceptual quality is influenced by a combination of geometric structure, color attributes, spatial organization, and multiscale distortion characteristics, rather than simple point-wise distances.

\textbf{PCQA Datasets and Benchmarks:} The development of PCQA has been strongly driven by subjective quality databases. Datasets such as SJTU-PCQA~\cite{sjtupcqa2020}, WPC~\cite{wpc2023}, SIAT-PCQD~\cite{wu2021siatpcqd}, BASICS~\cite{ak2023basics}, LS-PCQA~\cite{lspcqa2020}, and DPCD~\cite{liu2025dpcd} progressively expanded the scale, distortion diversity, and application scope of PCQA, covering static point clouds, immersive viewing conditions, large synthetic distortion sets, and dynamic point cloud sequences. Despite this diversity, most existing datasets are constructed by applying controlled distortions to reference point clouds and are annotated primarily with mean opinion scores (MOS).

\textbf{PCQA Methods:} Based on these benchmarks, no-reference PCQA methods have evolved along multiple technical directions, including projection-based pipelines, point-based and graph-based models, ranking-based learning, and self-supervised frameworks. Representative approaches include IT-PCQA~\cite{yang2022itpcqa}, GPA-Net~\cite{shan2023gpanet}, COPP-Net~\cite{cheng2023coppnet}, PRL-GQA~\cite{su2022prlgqa}, PKT-PCQA~\cite{liu2022pktpcqa}, and CoPA~\cite{shan2024copa}. More recently, multimodal and language-assisted methods, such as MM-PCQA~\cite{zhang2023mmpcqa}, D$^3$-PCQA~\cite{liu2025d3pcqa}, LMM-PCQA~\cite{zhang2024lmmpcqa}, CLIP-PCQA~\cite{liu2025clippcqa}, and PIT-QMM~\cite{gupta2025pitqmm}, have incorporated 2D projections, degradation descriptions, text supervision, and point–image–text fusion to enhance perceptual modeling.

\textbf{Evaluation Paradigm and Limitations:} Despite advances in datasets and model architectures, the evaluation paradigm of PCQA remains largely centered on scalar quality prediction, typically under synthetic distortion settings derived from reference point clouds. Consequently, existing benchmarks are effective at assessing how good a point cloud appears in terms of overall perceptual quality, but remain limited in evaluating what is wrong, how severe specific defects are, and whether a point cloud is usable for downstream tasks when facing heterogeneous, real-world data sources.

\subsection{Diagnostic and Interpretable Quality Assessment}

Recent quality assessment research in other modalities has increasingly explored interpretable and diagnostic evaluation beyond scalar score prediction. In video quality assessment, MaxVQA~\cite{wu2023maxvqa} introduces multi-factor annotations that relate overall quality to concrete perceptual attributes. In image quality assessment, benchmarks such as Q-Bench~\cite{qbench2024}, DepictQA~\cite{you2024depictqa}, DepictQA-Wild~\cite{you2024depictqawild}, and Grounding-IQA~\cite{chen2024groundingiqa} extend evaluation from single-score regression to low-level perception, distortion identification, comparative reasoning, and localized quality grounding. Similar trends are also observed in generative evaluation, where Q-Bench-Video~\cite{zhang2024qbenchvideo}, T$^3$Bench~\cite{he2023t3bench}, and MATE-3D~\cite{zhang2024mate3d} formulate quality assessment as a multi-dimensional problem involving perceptual fidelity, alignment, and consistency.

Collectively, these works reflect a broader shift in quality assessment from scalar regression toward structured quality cognition, where models are expected to diagnose factors, provide explanations, and support task-relevant reasoning. By contrast, although recent PCQA methods have begun to incorporate multimodal and language cues, existing PCQA benchmarks remain largely centered on overall quality scoring, offering limited support for diagnostic, explanatory, or decision-oriented evaluation.

\section{PointQ-Bench}
\label{sec:benchmark}


\subsection{Design Principles and Benchmark Scope}
\label{subsec:design_principles}

Inspired by recent low-level vision benchmarks such as Q-Bench and MedQ-Bench~\cite{qbench2024,medqbench2025}, which explicitly separate perception from reasoning, we define PointQ-Bench as a benchmark for multidimensional point-cloud quality understanding rather than a dataset used only for score regression.

Formally, PointQ-Bench is defined as
\begin{equation}
\mathcal{B}=\{(x_i,y_i^{mos},y_i^{lvl},Y_i^{iss},t_i)\}_{i=1}^{N},
\end{equation}
where $x_i$ is a point-cloud sample, $y_i^{mos}\in[1,5]$ is a continuous quality score, $y_i^{lvl}\in\{good,usable,bad\}$ is a three-level usability label, $Y_i^{iss}\subseteq\mathcal{S}$ is a multi-label defect set, and $t_i$ is an evidence-grounded text description. We follow four design principles: (1) mechanism-oriented coverage, whereby samples are organized by underlying degradation mechanisms rather than by source dataset;
(2) beyond-scalar supervision, which jointly evaluates quality gating, defect diagnosis, and explanatory reasoning, in line with the multidimensional perception-and-reasoning paradigm of recent benchmarks~\cite{qbench2024,medqbench2025};
(3) ambiguity preservation, which treats annotator disagreement near decision boundaries as informative uncertainty rather than noise; and
(4) information-density priority, which privileges diagnostically rich samples over indiscriminate dataset expansion with redundant or homogeneous instances.

\subsection{Data Sources and Integration}
\label{subsec:data_sources}

PointQ-Bench contains 3,083 samples drawn from seven public sources: 1,168 authentic, 1,315 simulated, and 600 AI-generated. We organize the benchmark around three complementary data regimes with distinct quality-degradation mechanisms: authentic scans capture real acquisition artifacts, simulated distortions provide controlled perturbations with clearer factor isolation, and AI-generated content introduces failure modes specific to modern generative pipelines. This design ensures that quality judgment is tested beyond a single narrow regime; detailed sampling protocols and composition statistics are provided in the Appendix.

The authentic branch captures quality issues from real acquisition and reconstruction. It consists of 369 samples from LiDAR-Net~\cite{lidarnet2024}, 300 from ScanObjectNN~\cite{scanobjectnn2019}, and 499 reconstructed from real multi-view videos in CO3D~\cite{co3d2021}, which complement each other in acquisition setting and object granularity. LiDAR-Net represents real indoor scene scans and naturally contains artifacts such as non-uniform density, occlusions, and scanning holes. ScanObjectNN contributes real object-centric scans; we use its no-background object-only test split so that the benchmark emphasizes intrinsic object quality rather than background clutter. To avoid excessive semantic overlap while preserving structural and material diversity, we retain a representative subset of 15 of CO3D's 50 categories.

The simulated branch provides controlled distortions that disentangle defect mechanisms from source-content variability. It includes 710 samples from ModelNet40-C~\cite{modelnet40c2022}, 344 samples from SJTU-PCQA~\cite{sjtupcqa2020}, and 261 samples from WPC~\cite{wpc2023}. ModelNet40-C serves as the primary controlled-distortion source because its clean CAD references reduce source-geometry confounds and make corruption effects easier to isolate; we exclude rigid rotation because it alters pose rather than intrinsic visual quality. SJTU-PCQA provides subjective quality annotations under controlled reference--distortion settings, while WPC supplements the branch with more practically relevant degradations such as downsampling, Gaussian noise, and compression artifacts. Together, these sources support evaluation under distortion regimes that are both analyzable and perceptually meaningful.

The AI-generated branch targets failure modes introduced by contemporary Text-to-3D systems. It is built on T23D-CompBench~\cite{t23dcompbench2025}, whose original release contains 3,600 textured meshes produced by ten state-of-the-art generators. From this source, we retain 600 samples corresponding to 75 prompts across eight generators after filtering out scene-level cases and samples with multiple objects distributed too far apart. This subset preserves representative generation failures in geometry and appearance while keeping the benchmark efficient for systematic evaluation.

\subsection{Annotation Protocol}
\label{subsec:annotation_protocol}

As illustrated in Figure~\ref{fig:teaser}, each sample was independently assessed by eight trained experts using both 2D projections and interactive 3D views. Here, ``expert'' refers to annotators with prior experience in 3D vision, point-cloud processing, or perceptual quality inspection. 

Following our unified protocol, each annotator $  r  $ recorded four raw fields for sample $  i  $: a continuous quality score $  x_{ir}\in\{1,2,3,4,5\}  $, a three-level usability label $  l_{ir}\in\{good,usable,bad\}  $, a multi-label impairment-source set $  Y_{ir}^{\mathrm{iss}}\subseteq\mathcal{S}  $, and a short evidence-grounded comment $  t_{ir}  $, where $  \mathcal{S}  $ denotes the fixed vocabulary of eight impairment types.

To mitigate annotator-specific leniency or severity bias, we aggregate the continuous scores via per-rater normalization and derive the final ordinal usability level by majority voting~\cite{bt5002023, p9102023}. After rigorous validity audits, we adopt a deliberate decision split: high-consensus samples form the official-clean split, while statistically ambiguous yet valid samples are retained as a separate boundary-valid split rather than discarded. This design preserves genuinely difficult cases near the usability transition boundary, enabling evaluation of calibration under realistic human disagreement instead of over-simplifying the test set into only high-consensus examples. Comprehensive details on rater training, impairment definitions and examples, calibration equations, audit rules, uncertainty thresholds, and split-wise statistics are provided in the Appendix.

\subsection{Perception Tasks and Evaluation Protocol}
\label{subsec:perception}

We decompose point-cloud quality perception into three progressive subtasks: anomaly sensing (\textit{Yes/No}), defect diagnosis (\textit{What}), and usability grading (\textit{How}). At the same time, considering that models may exhibit preference toward specific question phrasings, we provide three semantically equivalent prompt variants for each subtask and distribute them uniformly across evaluation samples.

\textit{Yes/No} evaluates a model's ability to sense noticeable quality degradation and uses accuracy as the primary metric. \textit{What} evaluates a model's ability to identify specific defect types, and is treated as a multi-label diagnosis task defined over $\mathcal{S}$. We use sample-level F1 as the primary metric, where \textit{NONE} is treated as an empty defect set rather than a regular class. \textit{How} evaluates a model's ability to make the final usability decision. We use macro-F1 as the primary metric, so as to better characterize model performance in the ambiguous intermediate region usable.

During inference, considering that strict answer-format constraints may partially suppress the model’s genuine perceptual ability, we do not require the tested models to follow a fixed output format. Instead, they are allowed to respond freely. We then use an LLM to normalize and parse the raw responses into the predefined canonical label space, after which the corresponding metrics are computed by a deterministic script.

\subsection{Open-ended Quality Reasoning and SSFRQ-5D Evaluation Protocol}
\label{subsec:reasoning}

Open-ended quality reasoning is subjective and linguistically complex, making traditional text-overlap metrics inadequate for reliable evaluation. To assess free-form quality reports structurally and diagnostically, we introduce SSFRQ-5D, a multi-dimensional evaluation protocol that compares model outputs against expert references.

Given a model output $\mathcal{O}$ and the corresponding expert reference text $\mathcal{R}$, each sample $i$ is assigned a discrete score vector
$\mathbf{s}_i=(s_i^{S1}, s_i^{S2}, s_i^{F}, s_i^{R}, s_i^{Q}) \in \{0,1,2\}^{5}$,
where 2, 1, and 0 indicate fully satisfied, partially satisfied, and seriously violated, respectively. Specifically, SSFRQ-5D evaluates open-ended quality reports along the following five complementary dimensions:

(1) Structural Sufficiency ($S1$) measures structural coverage. It requires $\mathcal{O}$ to include three basic elements: an overall quality summary, defect enumeration, and a final judgment, without requiring wording-level matching.
(2) Specificity ($S2$) measures evidential density. It penalizes vague and templated responses, and requires $\mathcal{O}$ to provide concrete low-level 3D geometric or texture cues, such as density abnormalities or topological breakage.
(3) Faithfulness ($F$) measures external consistency. By penalizing semantic contradictions and hallucinated defects, it ensures that $\mathcal{O}$ remains faithful to the observed facts in the reference text $\mathcal{R}$.
(4) Reasoning Coherence ($R$) measures internal consistency. It evaluates whether the chain of defect evidence constructed by the model in $\mathcal{O}$ is logically sufficient to support its final usability conclusion.
(5) Quality Accuracy ($Q$) independently evaluates the correctness of the final decision. Based on the absolute ordinal relation ($bad < usable < good$), it assigns discrete scores according to the level distance between $\mathcal{O}$ and $\mathcal{R}$.

In particular, $S1$ and $S2$ distinguish structural completeness from evidential specificity, while $F$ and $R$ separate consistency with the reference from internal logical consistency.

\subsection{Automatic Judges and Human--AI Agreement}

\label{subsec:judge_alignment}

To ensure the reliability and validity of the automated evaluation pipeline, we validate the objective and subjective components separately. For the objective perception tasks, we manually verify the parser outputs on 400 randomly sampled cases to assess whether the answer extraction process introduces measurable errors into evaluation.

For the more subjective SSFRQ-5D reasoning evaluation, we assess the agreement between the automatic judge and expert annotations. Specifically, we randomly sample 200 reasoning cases and ask two senior 3D vision experts to independently score all five dimensions under a double-blind protocol. After pooling the two experts, each dimension contains 400 human judgments. Following prior protocols for ordinal agreement analysis, we report both exact agreement (EA) and quadratic weighted Cohen's kappa ($\kappa_w$)~\cite{cohen1968weighted,medqbench2025}. The exact agreement is defined as
\begin{equation}
\mathrm{EA}=\frac{1}{N}\sum_{n=1}^{N}\mathbb{I}(\hat{y}_n=y_n),
\end{equation}
where $\hat{y}_n$ denotes the automatic score and $y_n$ denotes the expert score for the $n$-th judgment. To further account for the ordinal nature of the three-level scale, we compute quadratic weighted Cohen's kappa:
\begin{equation}
\kappa_w = 1-\frac{\sum_{i,j} w_{ij}O_{ij}}{\sum_{i,j} w_{ij}E_{ij}},\qquad
w_{ij}=\frac{(i-j)^2}{(k-1)^2},
\label{eq:qwk}
\end{equation}
where $O_{ij}$ and $E_{ij}$ denote the observed and expected agreement matrices, respectively, and $k=3$ is the number of ordinal levels~\cite{cohen1968weighted}. Final alignment results are reported in Appendix.

\input{tables/main_table}

\section{Results on PointQ-Bench}
To systematically evaluate point-cloud quality understanding, we benchmark 14 representative models under a unified zero-shot setting. These span three categories: proprietary 2D MLLMs (GPT-5~\cite{singh2025openaigpt5card}, GPT-5-mini~\cite{singh2025openaigpt5card}, Claude-4.5-haiku~\cite{anthropic2025claudehaiku45}), open-source 2D MLLMs (Qwen-3.5~\cite{qwen35blog}, InternVL-3.5~\cite{wang2025internvl35advancingopensourcemultimodal}, Mistral-Small-3.2~\cite{MistralAI}, Gemma-3~\cite{gemmateam2025gemma3technicalreport}), and native 3D VLMs (PointLLM~\cite{xu2024pointllm}, ShapeLLM~\cite{zekunqi2024shapellm}, MiniGPT-3D~\cite{minigpt}).

For 2D MLLMs, we introduce varying input configurations—\textbf{sv1} (single-view), \textbf{mv2} (two-view), \textbf{mv4} (four-view), and \textbf{mv6} (six-view) projections—to assess spatial context utilization. To ensure fair comparison against 3D VLMs, which lack configurable reasoning modes, 2D MLLMs are evaluated under no-thinking or low-thinking settings. Unless specified, all models use deterministic decoding (temperature set to 0).

The ``Junior-level Human'' baseline in Table~\ref{tab:main} is obtained from two undergraduate participants who had no prior background in point clouds and received no task-specific training before evaluation. We intentionally include this baseline as a lightweight novice-human reference, rather than an expert ceiling, to reflect how non-specialist users perform under minimal prior knowledge. Human participants were evaluated using the same task instructions and the same visual inputs as the corresponding model setting.

\subsection{Perception Tasks}
Table~\ref{tab:main} reports the results of MLLMs and the junior-level human baseline on the three perception tasks of point-cloud quality assessment, with a random-guess baseline for reference.

\textbf{Baseline Comparisons.} Compared with random guessing, leading 2D MLLMs improve Yes/No from 50.67\% to 88.68–89.00\%, What from 14.70\% to 42.32\%, and How from 34.26\% to 50.13\%, indicating that their predictions are not merely driven by label priors. Compared with the junior-level human baseline (86.25/36.67/48.35), proprietary 2D MLLMs are already competitive on fine-grained perception: GPT-5~\cite{singh2025openaigpt5card} achieves the best What and How scores at 42.32\% and 50.13\%, exceeding the human baseline by +5.65 and +1.78 points, while Claude-4.5-haiku~\cite{anthropic2025claudehaiku45} reaches 88.68\% on Yes/No, slightly surpassing human performance by +2.43 points.

\textbf{Hierarchy Across Model Families.} Table~\ref{tab:main} shows a clear hierarchy across model families. Proprietary 2D MLLMs achieve the most balanced perception performance, with GPT-5 leading on \textit{What} (42.32\%) and \textit{How} (50.13\%), and Claude-4.5-haiku achieving the best \textit{Yes/No} accuracy (88.68\%).  Open-source 2D MLLMs form a strong second tier and consistently outperform most native 3D VLMs. Their strengths are also task-dependent: Qwen-3.5 (9B) performs better on \textit{What} (39.95\%), whereas Qwen-3.5 (27B) performs better on \textit{How} (46.05\%). This pattern suggests that scaling up model size does not necessarily yield uniformly better point-cloud quality understanding.

\textbf{Performance of 3D Vision-Language Models.} In contrast, current 3D VLMs show only limited competitiveness. ShapeLLM (13B) is the only 3D model that reaches top-tier performance on \textit{Yes/No} accuracy (89.00\%), even slightly surpassing Claude-4.5-haiku on coarse issue detection. However, this advantage does not extend to the more fine-grained tasks. On both \textit{What} and \textit{How}, all 3D VLMs remain clearly behind the strongest 2D MLLMs, while PointLLM and MiniGPT-3D perform substantially worse overall.

Overall, these results reveal two main findings: (1) current MLLMs are already capable of coarse-grained defect perception, but fine-grained diagnosis and quality grading remain significantly more challenging; and  (2) current native 3D VLMs still fall short of strong 2D MLLMs on robust quality understanding, suggesting that native 3D input has not yet been effectively translated into fine-grained quality reasoning.

\subsection{Cognition Tasks}
Table~\ref{tab:main} reveals that the main bottleneck of point-cloud quality cognition lies in grounded reasoning rather than report fluency. Strong 2D MLLMs already achieve near-ceiling scores on structural dimensions such as S1, S2, and R, but remain clearly weaker on Faithfulness (F) and Quality Accuracy (Q). For example, GPT-5 reaches 2.00/2.00/1.95 on S1/S2/R, yet only 1.07/1.39 on F/Q, indicating that well-structured reports do not necessarily reflect faithful defect grounding or accurate final judgments. More importantly, this gap is not an isolated case of GPT-5~\cite{singh2025openaigpt5card}: the same high-S1/S2 but low-F/Q pattern also appears in GPT-5-mini~\cite{singh2025openaigpt5card} and Claude-4.5-haiku~\cite{anthropic2025claudehaiku45}, suggesting a shared limitation in evidence grounding rather than report organization.

This gap is not closed by 3D VLMs. Although some 3D models remain competitive on structural coherence, they still lag substantially behind strong 2D MLLMs on Faithfulness and overall reasoning quality. For instance, ShapeLLM-13B~\cite{zekunqi2024shapellm} achieves a relatively high R score of 1.99, but only 0.44 on F and 6.32 on SSFRQ-5D, compared with 1.07 and 8.41 for GPT-5. Overall, current models are much better at producing plausible quality reports than at grounding them in reliable defect evidence and calibrated usability decisions.

\subsection{Analysis of Multi-View Inputs for 2D VLMs}
\begin{figure}[htbp]
    \centering

    \begin{subfigure}[b]{0.50\textwidth}
        \centering
        \includegraphics[width=\textwidth]{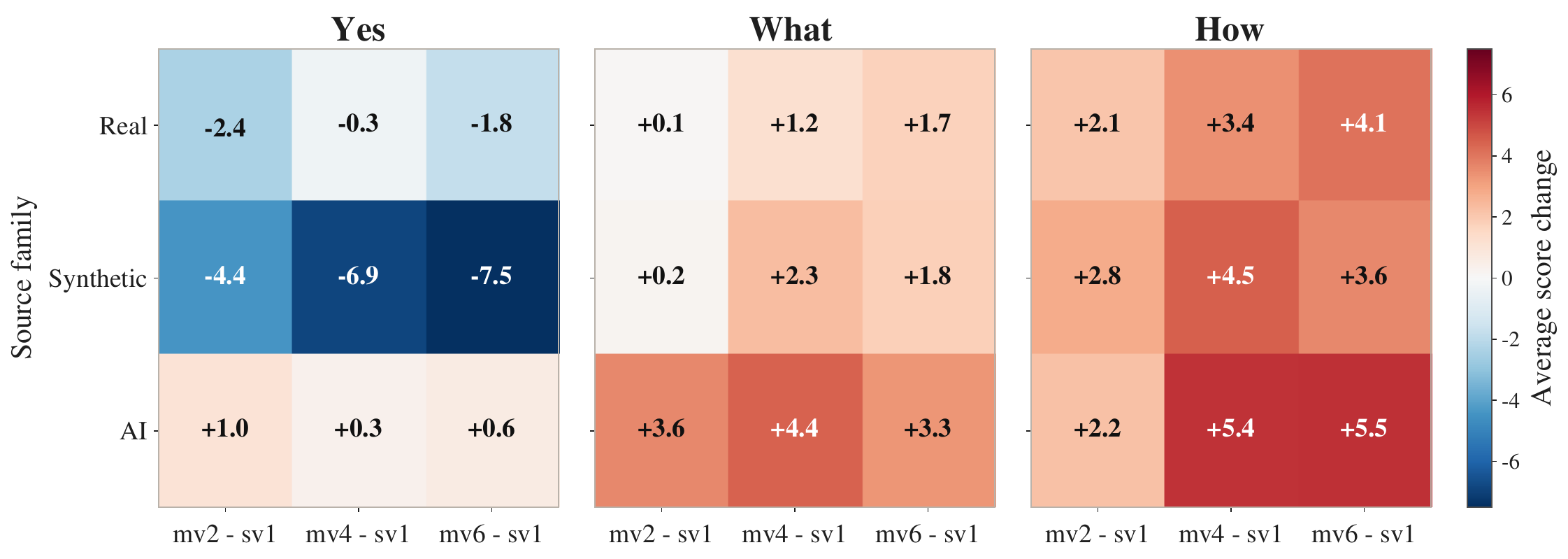}
        \caption{Average multi-view gain relative to sv1 across different source families and tasks.}
        \label{fig:sub_a}
    \end{subfigure}
    \hfill
    \begin{subfigure}[b]{0.48\textwidth}
        \centering
        \includegraphics[width=\textwidth]{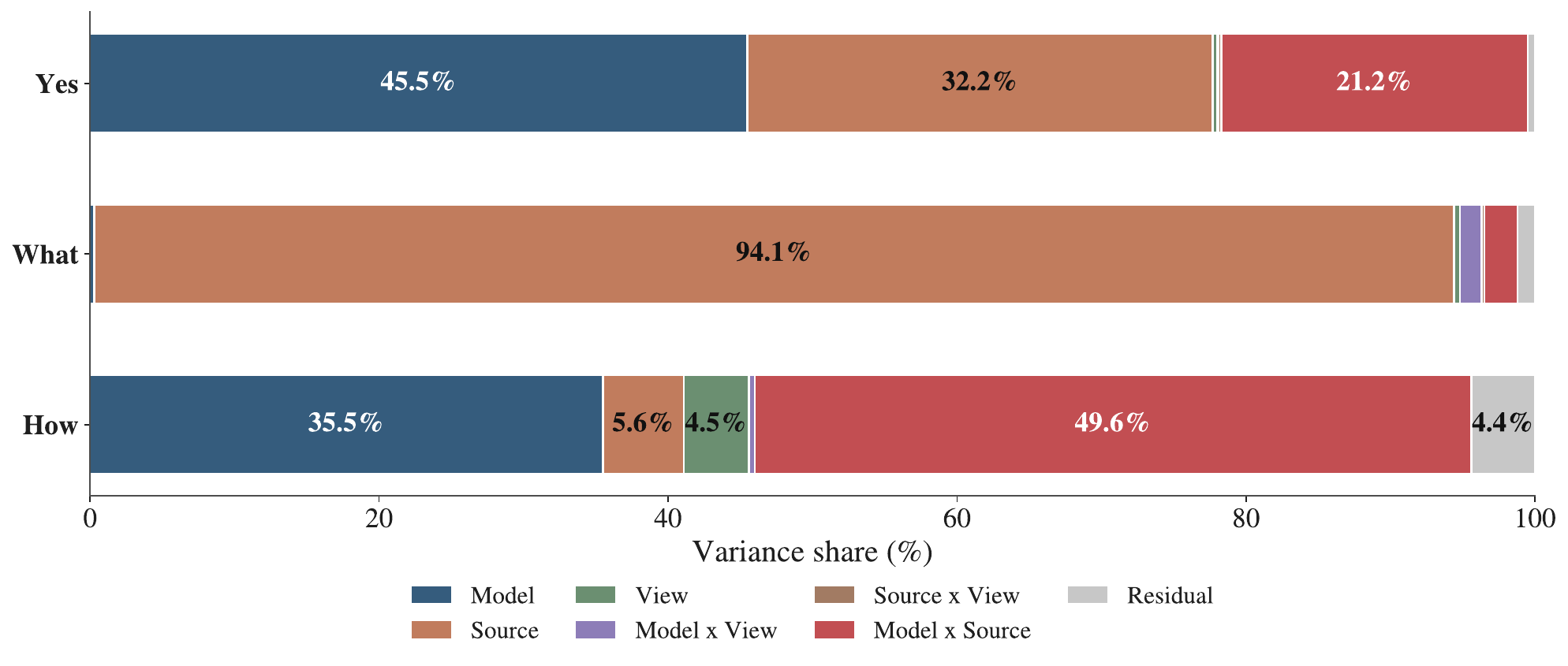}
        \caption{Descriptive variance decomposition across factors.}
        \label{fig:sub_b}
    \end{subfigure}
    
    \vspace{1em} 
    
    \begin{subfigure}[b]{0.50\textwidth}
        \centering
        \includegraphics[width=\textwidth]{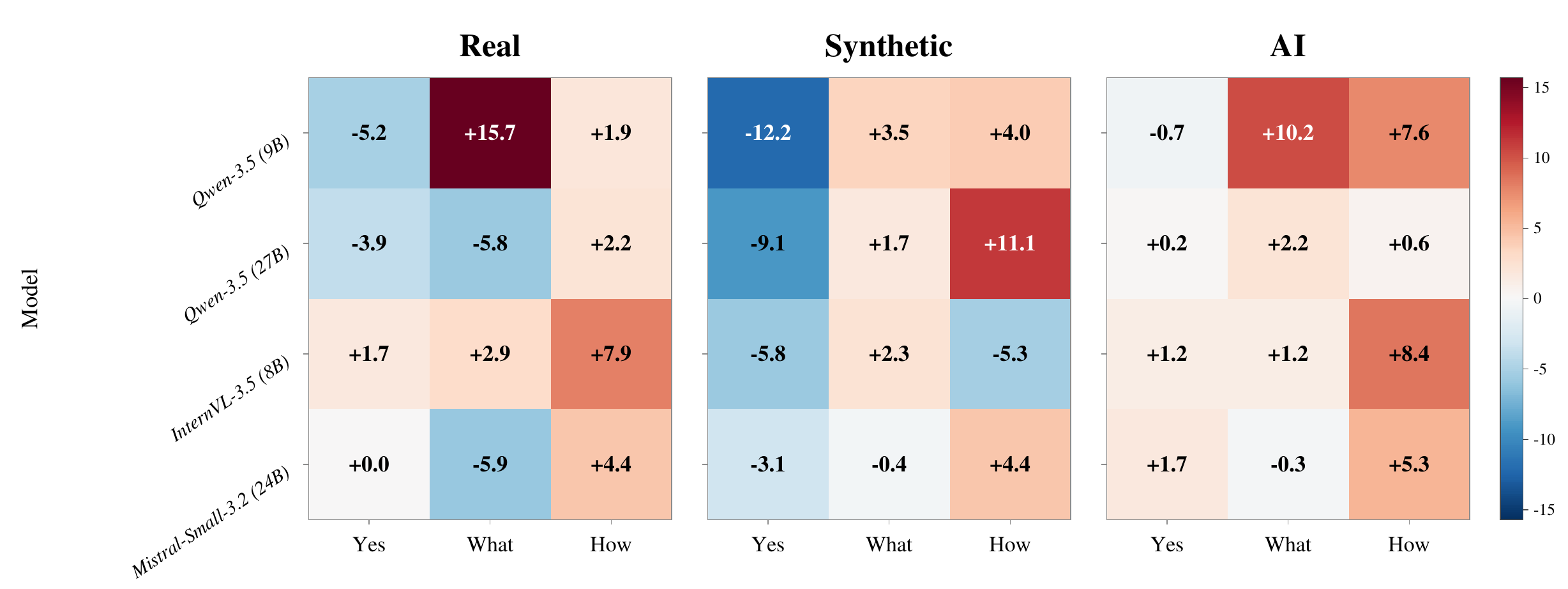}
        \caption{Multi-view effect (mv6 relative to sv1) detailed by specific models.}
        \label{fig:sub_c}
    \end{subfigure}
    \hfill
    \begin{subfigure}[b]{0.48\textwidth}
        \centering
        \includegraphics[width=\textwidth]{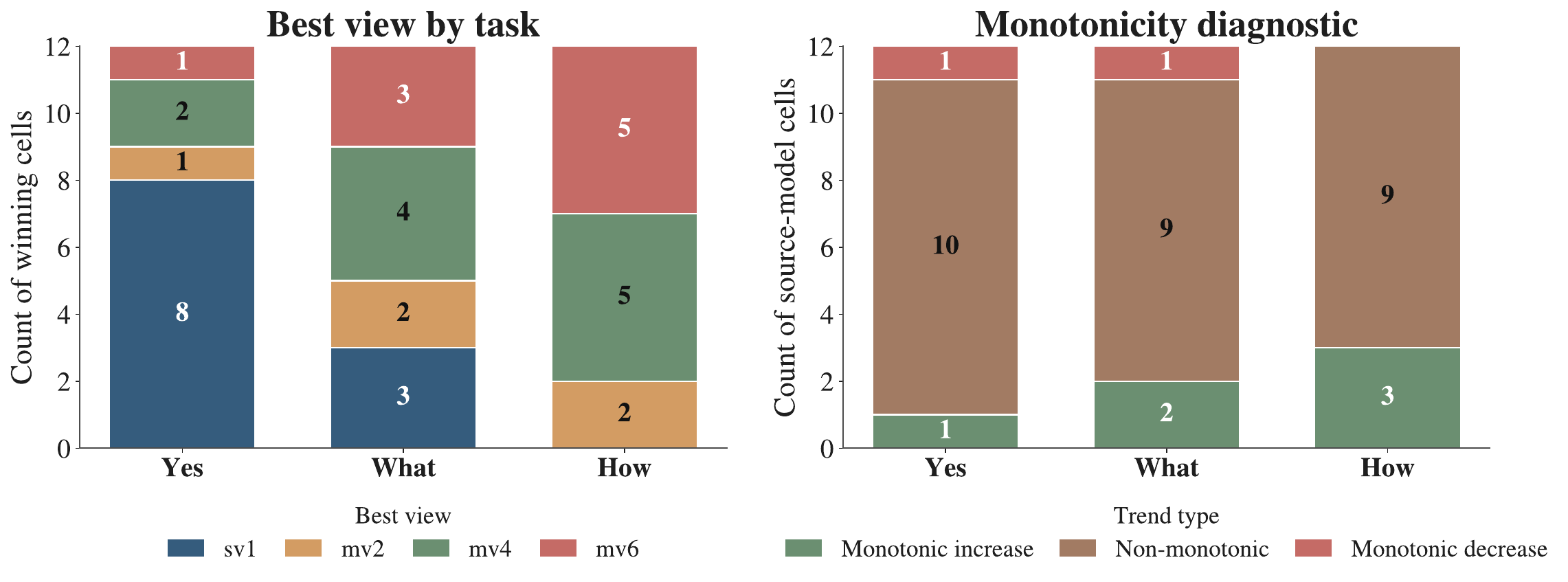}
        \caption{Monotonicity and best-view winning counts per task.}
        \label{fig:sub_d}
    \end{subfigure}

  \vspace{-1em}
    \caption{Comprehensive analysis of multi-view effects. (a) Average gains across source families; (b) variance decomposition; (c) final multi-view effects across models; and (d) monotonicity diagnostics and best view counts.}
\label{fig:multi_view}
\end{figure}

Figure~\ref{fig:multi_view} reports a multi-view ablation on four
representative open-source 2D VLMs, showing that the effect of
additional views depends more on task, source family, and model
than on view count.

A clear task-dependent pattern emerges in Figure~\ref{fig:multi_view}(a). Multi-view inputs consistently benefit usability grading (\textit{How}), with gains reaching +4.1 on Real, +4.5 on Synthetic, and +5.5 on AI-generated data under mv6 relative to sv1. The gains on defect diagnosis (\textit{What}) are smaller but mostly positive, while \textit{Yes/No} often degrades on Real and Synthetic data (e.g., \(-2.4\), \(-4.4\) under mv2, and \(-1.8\), \(-7.5\) under mv6), indicating that additional views do not reliably improve coarse defect sensing.

Variance analysis in Figure~\ref{fig:multi_view}(b) confirms this
limited contribution: the main effect of view count explains only
0.20\%, 0.45\%, and 4.55\% of the variance for
\textit{Yes/No}, \textit{What}, and \textit{How}, respectively, far
below the contributions of source, model, and their interactions.
Figures~\ref{fig:multi_view}(c)--(d) further show strong
source--model--task heterogeneity and that the best-performing
setting is not consistently associated with more views. Overall,
multi-view prompting helps only in specific regimes.

\input{tables/3d}

\subsection{Analysis of Native 3D VLMs}
\label{subsec:3d_ablations}

To understand the limits of native 3D quality understanding, we perform ablations on point budget and sampling strategy for the more competitive 3D VLMs (Table~\ref{tab:main}). Because ModelNet40-C and ScanObjectNN lack sufficient native points for all tested budgets, Table~\ref{tab:sampling_sensitivity} uses the remaining common 2,073-sample subset. The results reveal a non-monotonic pattern that echoes the 2D multi-view study: increasing the number of input points does not consistently improve coarse anomaly sensing (\textit{Yes/No}), whereas fine-grained defect identification (\textit{What}) and usability grading (\textit{How}) generally benefit from richer geometric input.

This task-dependent pattern suggests that additional 3D evidence is not uniformly beneficial: extra points can introduce redundant or conflicting cues for \textit{Yes/No}, whereas \textit{What} and \textit{How} rely more on localized and complementary defect evidence. At the same point budget, random sampling (RS) frequently outperforms farthest point sampling (FPS) on \textit{What} and \textit{How}, suggesting that uniform global coverage can dilute localized defect clusters relevant to diagnosis.

\input{tables/pcaqa}

\subsection{Analysis of Conventional PCQA Models}
We further evaluate representative no-reference PCQA models in the conventional scalar prediction setting. Since none is fine-tuned on PointQ-Bench, Table~\ref{tab:pcqa-aggregated} serves as a zero-shot cross-source transfer reference rather than an architecture ranking or an upper bound. By retaining MOS annotations while covering simulated, authentic, and AI-generated data, PointQ-Bench places conventional scalar PCQA in a broader and more heterogeneous evaluation setting.

These models preserve some effectiveness on the \textit{Simulated} branch but transfer poorly to \textit{Authentic} and \textit{AI-generated} data, where rank correlations weaken and RMSE increases. This pattern suggests that the key limitation is cross-source generalization rather than scalar supervision itself: models trained mainly on synthetic or compression-oriented distributions remain more effective in familiar regimes but struggle with real scanning artifacts and generative failures.

PointQ-Bench therefore serves a dual role: it supports conventional MOS prediction while exposing where source-specialized scalar models become unreliable, motivating broader quality understanding beyond scalar scores.

\subsection{Usability Grading Across Consensus Strata}
\label{subsec:how_consensus}

\begin{figure}[t]
  \centering
 
  \includegraphics[width=\columnwidth]{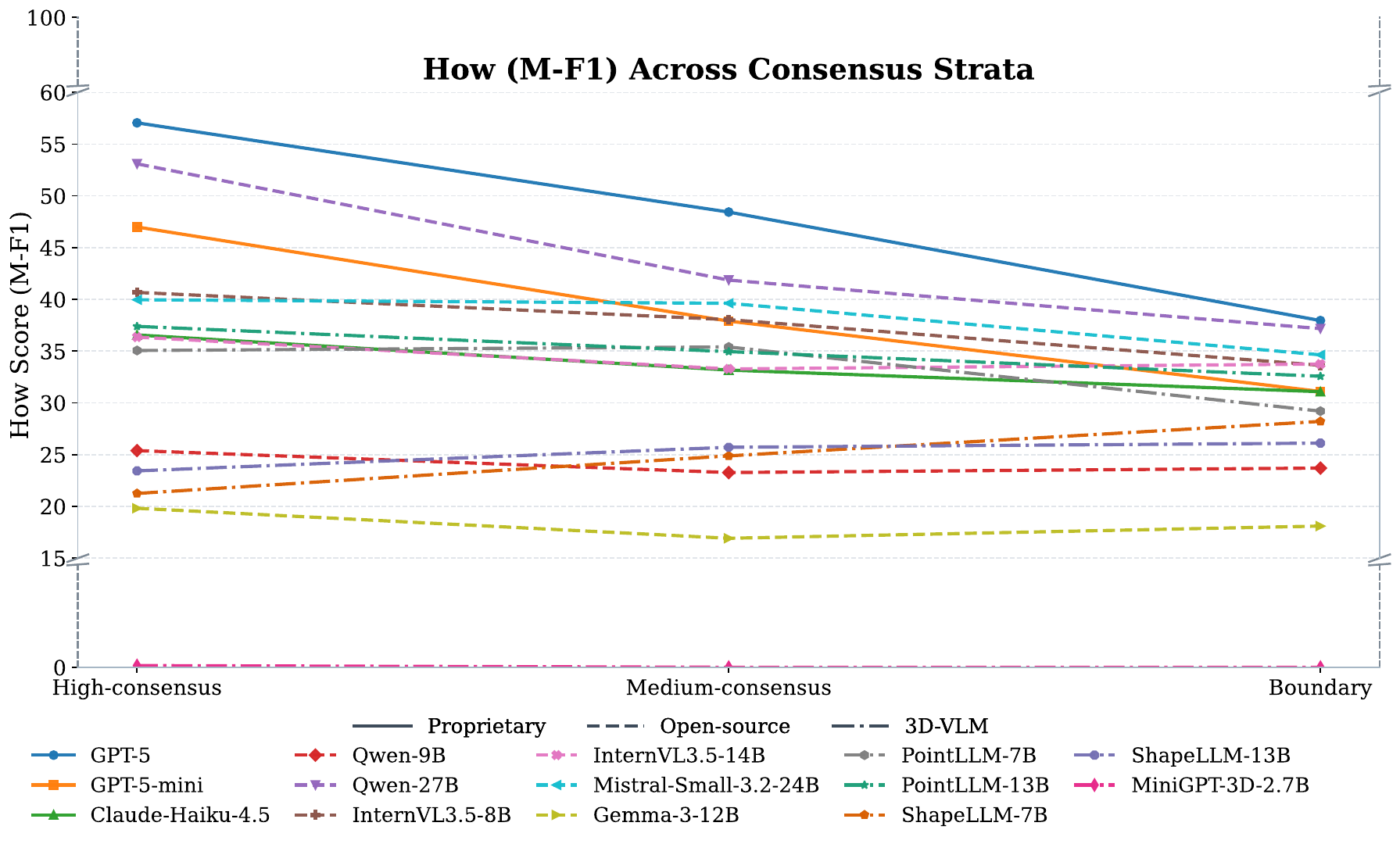}
   \vspace{-1em}
  \caption{Usability grading across consensus strata.}
\label{fig:consensus_trend}
\end{figure}

Figure~\ref{fig:consensus_trend} reports usability grading (\textit{How} M-F1) across three consensus strata. The \textit{Boundary} split captures the ambiguous usable transition band ($\text{MOS} \in [2.0, 3.5]$) rather than annotation noise.

\textbf{The Usability Transition Band as a Calibration Bottleneck.} As human consensus decreases from \textit{High} to \textit{Boundary}, the \textit{How} scores of all competitive models suffer severe degradation. For instance, GPT-5~\cite{singh2025openaigpt5card} drops from 57.05 to 37.95, and Qwen-27B~\cite{qwen35blog} drops from 53.09 to 37.17. Even the strongest systems fail to surpass an M-F1 of 40 in the boundary zone. This confirms that the primary bottleneck of quality grading is not coarse defect sensing, but rather the precise calibration of severity thresholds in ambiguous transition bands. 

\textbf{Boundary Samples Act as a Capability Equalizer.} A striking phenomenon is the sharp compression of the leaderboard in the ambiguous zone. On \textit{High-consensus} samples, the score gap between the top five models is approximately 17.1 points; however, on \textit{Boundary} samples, this gap collapses to merely 4.32 points. This gap compression suggests that frontier-model advantages narrow in the usable transition zone, where human experts also disagree and calibration becomes more difficult.

\begin{figure}[t]
  \centering
  \includegraphics[width=1\columnwidth]{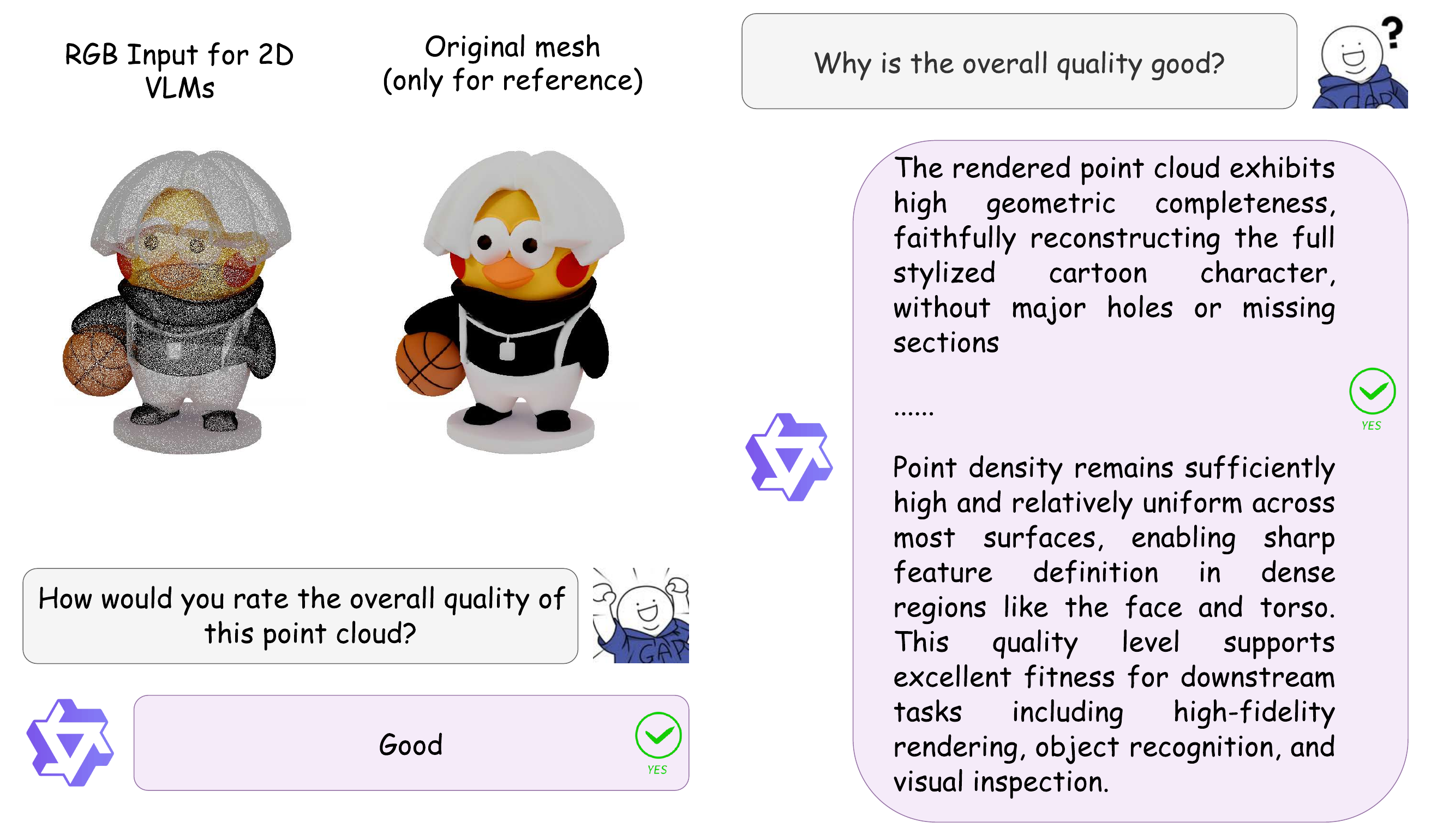}
  \label{show}
  \caption{Demonstrations of PointQ-Bench.}
\end{figure}

\textbf{False Robustness and Objective Mismatch.} The slope of the curves further exposes distinct inductive biases. The flat trajectories of models like Gemma-3~\cite{gemmateam2025gemma3technicalreport} do not imply robustness; rather, they reflect a low sensitivity to ordinal quality scales. More alarmingly, models like ShapeLLM~\cite{zekunqi2024shapellm} exhibit a paradoxical rise in performance on \textit{Boundary} samples (e.g., 23.43 $\rightarrow$ 26.11). This should not be interpreted as boundary mastery. Instead, it exposes a severe prediction collapse toward the central usable class. Such behaviors stem from a fundamental training-objective mismatch: while current 2D MLLMs and 3D VLMs are heavily optimized for general visual question answering, object semantics, or spatial captioning, they are rarely trained to handle severity mapping or threshold calibration under human uncertainty. Consequently, general vision-language priors help resolve obvious cases but fail to automatically produce boundary-aware quality calibration.

\FloatBarrier
\section{Conclusion}
\label{sec:conclusion}
In this work, PointQ-Bench extends point cloud quality assessment beyond scalar prediction toward grounded diagnosis and usability calibration across heterogeneous sources and ambiguous quality boundaries. Extensive evaluation and agreement-conditioned analyses show that, under the evaluated protocols, current systems are stronger at coarse anomaly sensing than at faithful diagnosis and boundary-aware grading, with the latter degrading sharply on ambiguous samples. Multi-view and point-budget analyses further show that additional evidence helps when it provides complementary, localized defect cues; otherwise, it may introduce decision noise. These findings motivate grounded defect diagnosis, uncertainty-aware calibration, and source-robust quality reasoning rather than simply scaling input evidence. Overall, PointQ-Bench exposes the gap between plausible quality descriptions and reliable quality decisions and provides a diagnostic testbed for more interpretable point-cloud quality understanding.

\section*{Acknowledgments}
This work was supported by the National Natural Science Foundation
of China under Grant No.~42571444. The authors thank the annotators
for their efforts in data annotation and the domain experts for their
assistance with protocol validation.

\bibliographystyle{unsrtnat}
\bibliography{references}

\end{document}

%% file: tables/main_table.tex
\begin{table}[t]
\centering
\captionsetup{skip=1pt}
\caption{Performance of different models on the perception and cognition tasks of PointQ-Bench. First place in each column is \textbf{bolded}, and second places are \uline{underlined}. Random guess and junior-level human are excluded from ranking.}
\renewcommand{\arraystretch}{1.05} 

\label{tab:main}
\resizebox{\textwidth}{!}{%
\begin{tabular}{lccc|cccccc}
\toprule
\textbf{Sub-categories} &
\multicolumn{3}{c|}{\textbf{Perception}} &
\multicolumn{6}{c}{\textbf{Cognition}} \\
\cmidrule(lr){2-4}\cmidrule(lr){5-10}
\textbf{Model} & \textbf{yesno}$_{acc. (\%)}$$\uparrow$ & \textbf{What}$_{sample\_f1(\%)}$$\uparrow$ & \textbf{How}$_{macro\_f1(\%)}$$\uparrow$
& \textbf{S1/Str.$\uparrow$} & \textbf{S2/Spec.$\uparrow$} & \textbf{F/Faith.$\uparrow$} & \textbf{R/Coh.$\uparrow$} & \textbf{Q/Acc.$\uparrow$} & \textbf{SSFRQ-5D$_{10}\uparrow$} \\
\midrule

\rowcolor{mygreen}
\multicolumn{10}{c}{\textbf{Human (Point Cloud + Image) \& Random Guess }} \\
Junior-level Human  & 86.25 & 36.67 & 48.35 & - & - & - & - & - & - \\
Random guess        & 50.67 & 14.70 & 34.26 & - & - & - & - & - & - \\
\hline

\rowcolor{myorange}
\multicolumn{10}{c}{\textbf{\textit{Proprietary Large Vision-Language Models (mv6)}}} \\
GPT-5~\cite{singh2025openaigpt5card}              & 83.59 & \textbf{42.32} & \textbf{50.13} & \textbf{2.00} & \textbf{2.00} & \textbf{1.07} & 1.95 & \uline{1.39} & \textbf{8.41} \\
GPT-5-mini~\cite{singh2025openaigpt5card}         & 82.33 & 39.60 & 40.91 & \textbf{2.00} & \textbf{2.00} & \uline{0.99} & 1.78 & \textbf{1.44} & \uline{8.21} \\
Claude-4.5-haiku~\cite{anthropic2025claudehaiku45}   & \uline{88.68} & 36.88 & 34.40 & \textbf{2.00} & \textbf{2.00} & 0.67 & 1.71 & 1.38 & 7.76 \\
\hline

\rowcolor{myyellow}
\multicolumn{10}{c}{\textbf{\textit{Open-Source Large Vision-Language Models (mv6)}}} \\
Qwen-3.5 (9B)~\cite{qwen35blog}           & 78.27 & \uline{39.95} & 24.68 & \textbf{2.00} & 1.93 & 0.81 & 1.97 & 1.18 & 7.89 \\
Qwen-3.5 (27B)  ~\cite{qwen35blog}        & 82.45 & 33.87 & \uline{46.05} & \textbf{2.00} & \uline{1.99} & 0.73 & \textbf{1.99} & 1.06 & 7.77 \\
InternVL-3.5 (8B) ~\cite{wang2025internvl35advancingopensourcemultimodal}      & 38.70 & 34.83 & 38.28 & 1.93 & 1.89 & 0.74 & 1.92 & 1.16 & 7.72 \\
InternVL-3.5 (14B) ~\cite{wang2025internvl35advancingopensourcemultimodal}     & 75.38 & 30.24 & 35.83 & \textbf{2.00} & 1.72 & 0.73 & \uline{1.98} & 1.26 & 7.76 \\
Mistral-Small-3.2 (24B)~\cite{MistralAI} & 87.87 & 34.30 & 38.56 & \textbf{2.00} & 1.94 & 0.69 & 1.77 & 1.33 & 7.73 \\
gemma-3 (12B)    ~\cite{gemmateam2025gemma3technicalreport}       & \textbf{89.00} & 37.30 & 18.50 & \textbf{2.00} & 1.91 & 0.71 & 1.89 & 1.17 & 7.77 \\
\hline

\rowcolor{myblue}
\multicolumn{10}{c}{\textbf{\textit{3D Vision-Language Models (8192/Random Sample)}}} \\
PointLLM (7B) ~\cite{xu2024pointllm}      & 11.11 & 28.08 & 34.08 & 1.06 & 0.36 & 0.29 & 1.80 & 0.81 & 4.31 \\
PointLLM (13B) ~\cite{xu2024pointllm}     & 11.19 & 14.95 & 35.64 & 1.13 & 0.74 & 0.15 & 1.78 & 0.69 & 4.49 \\
ShapeLLM (7B) ~\cite{zekunqi2024shapellm}      & 66.17 & 26.22 & 24.18 & 1.75 & 0.84 & 0.39 & \textbf{1.99} & 0.96 & 5.93 \\
ShapeLLM (13B)  ~\cite{zekunqi2024shapellm}    & \textbf{89.00} & 32.42 & 25.01 & \uline{1.94} & 0.98 & 0.44 & \textbf{1.99} & 0.97 & 6.32 \\
MiniGPT-3D (2.7B) ~\cite{minigpt}  & 59.06 & 10.96 & 0.16  & 0.36 & 0.16 & 0.43 & 0.90 & 1.16 & 3.00 \\

\bottomrule
\end{tabular}%
}
\end{table}

%% file: tables/3d.tex
\begin{table}[t]
\centering
\setlength{\tabcolsep}{4.5pt} 
\captionsetup{skip=1pt}
\renewcommand{\arraystretch}{1.15} 
\caption{Sensitivity to point density and sampling strategy.}
\label{tab:sampling_sensitivity}

\newcommand{\up}[1]{\textcolor{red}{\scriptsize ($\uparrow$#1)}}
\newcommand{\dn}[1]{\textcolor{teal}{\scriptsize ($\downarrow$#1)}}

\resizebox{\columnwidth}{!}{%
\begin{tabular}{ll ccc}
\toprule
\textbf{Model} & \textbf{Setting}
  & \textbf{Yes-or-No}$\uparrow$
  & \textbf{What}$\uparrow$
  & \textbf{How}$\uparrow$ \\
\midrule

\multirow{4}{*}{\textbf{PointLLM(7B)~\cite{xu2024pointllm} }}
  & \cellcolor{baselinegray}\textbf{8192 / RS}
    & \cellcolor{baselinegray}15.78
    & \cellcolor{baselinegray}\textbf{30.66}
    & \cellcolor{baselinegray}\textbf{28.62} \\
  & 4096 / RS & \underline{15.95} \up{0.17} & 30.44 \dn{0.22} & 27.88 \dn{0.74} \\
  & 2048 / RS & 15.90 \up{0.12} & \underline{30.47} \dn{0.19} & 26.91 \dn{1.71} \\
  & 8192 / FPS& \textbf{16.03} \up{0.25} & 30.38 \dn{0.28} & \underline{28.10} \dn{0.52} \\
\midrule

\multirow{4}{*}{\textbf{PointLLM(13B)~\cite{xu2024pointllm} }}
  & \cellcolor{baselinegray}\textbf{8192 / RS}
    & \cellcolor{baselinegray}15.58
    & \cellcolor{baselinegray}\textbf{17.08}
    & \cellcolor{baselinegray}\underline{35.12} \\
  & 4096 / RS & 15.71 \up{0.13} & 16.62 \dn{0.46} & 33.74 \dn{1.38} \\
  & 2048 / RS & \underline{15.83} \up{0.25} & 16.54 \dn{0.54} & 33.01 \dn{2.11} \\
  & 8192 / FPS& \textbf{15.92} \up{0.34} & \underline{16.87} \dn{0.21} & \textbf{35.36} \up{0.24} \\
\midrule

\multirow{4}{*}{\textbf{ShapeLLM(7B)~\cite{zekunqi2024shapellm}}}
  & \cellcolor{baselinegray}\textbf{8192 / RS}
    & \cellcolor{baselinegray}66.18
    & \cellcolor{baselinegray}\textbf{29.24}
    & \cellcolor{baselinegray}\textbf{24.66} \\
  & 4096 / RS & \textbf{68.74} \up{2.56} & 28.02 \dn{1.22} & 22.18 \dn{2.48} \\
  & 2048 / RS & \underline{68.41} \up{2.23} & \underline{28.19} \dn{1.05} & 22.36 \dn{2.30} \\
  & 8192 / FPS& 67.81 \up{1.63} & 27.78 \dn{1.46} & \underline{23.54} \dn{1.12} \\
\midrule

\multirow{4}{*}{\textbf{ShapeLLM(13B)~\cite{zekunqi2024shapellm}}}
  & \cellcolor{baselinegray}\textbf{8192 / RS}
    & \cellcolor{baselinegray}83.92
    & \cellcolor{baselinegray}\textbf{34.56}
    & \cellcolor{baselinegray}\textbf{27.28} \\
  & 4096 / RS & \textbf{84.24} \up{0.32} & 34.21 \dn{0.35} & \underline{26.91} \dn{0.37} \\
  & 2048 / RS & \underline{84.18} \up{0.26} & 34.17 \dn{0.39} & 26.38 \dn{0.90} \\
  & 8192 / FPS& 84.06 \up{0.14} & \underline{34.47} \dn{0.09} & 26.74 \dn{0.54} \\
\bottomrule
\end{tabular}%
}
\end{table}

%% file: tables/pcaqa.tex
\begin{table}[t]
\centering
\caption{Zero-Shot Transfer of Conventional PCQA Models.}
\label{tab:pcqa-aggregated}

\resizebox{\textwidth}{!}{%
\small
\setlength{\tabcolsep}{6pt}
\renewcommand{\arraystretch}{1.15}

\begin{tabular}{
l
>{\columncolor{myorange!25}}c
>{\columncolor{myorange!25}}c
>{\columncolor{myorange!25}}c
>{\columncolor{myorange!25}}c
>{\columncolor{myyellow!25}}c
>{\columncolor{myyellow!25}}c
>{\columncolor{myyellow!25}}c
>{\columncolor{myyellow!25}}c
>{\columncolor{myblue!25}}c
>{\columncolor{myblue!25}}c
>{\columncolor{myblue!25}}c
>{\columncolor{myblue!25}}c
>{\columncolor{mygreen!25}}c
>{\columncolor{mygreen!25}}c
>{\columncolor{mygreen!25}}c
>{\columncolor{mygreen!25}}c
}

\toprule

\textbf{Dataset Type}
  & \multicolumn{4}{c}{\cellcolor{myorange}\textbf{Authentic}}
  & \multicolumn{4}{c}{\cellcolor{myyellow}\textbf{Simulated}}
  & \multicolumn{4}{c}{\cellcolor{myblue}\textbf{AI-Generated}}
  & \multicolumn{4}{c}{\cellcolor{mygreen}\textbf{Overall}} \\

\cmidrule(lr){2-5}
\cmidrule(lr){6-9}
\cmidrule(lr){10-13}
\cmidrule(lr){14-17}

\textbf{Model}
  & \cellcolor{white}\textbf{SRCC$\uparrow$}
  & \cellcolor{white}\textbf{PLCC$\uparrow$}
  & \cellcolor{white}\textbf{KRCC$\uparrow$}
  & \cellcolor{white}\textbf{RMSE$\downarrow$}

  & \cellcolor{white}\textbf{SRCC$\uparrow$}
  & \cellcolor{white}\textbf{PLCC$\uparrow$}
  & \cellcolor{white}\textbf{KRCC$\uparrow$}
  & \cellcolor{white}\textbf{RMSE$\downarrow$}

  & \cellcolor{white}\textbf{SRCC$\uparrow$}
  & \cellcolor{white}\textbf{PLCC$\uparrow$}
  & \cellcolor{white}\textbf{KRCC$\uparrow$}
  & \cellcolor{white}\textbf{RMSE$\downarrow$}

  & \cellcolor{white}\textbf{SRCC$\uparrow$}
  & \cellcolor{white}\textbf{PLCC$\uparrow$}
  & \cellcolor{white}\textbf{KRCC$\uparrow$}
  & \cellcolor{white}\textbf{RMSE$\downarrow$} \\

\midrule

ResSCNN~\cite{lspcqa2020}
  & -0.141 & 0.202 & -0.097 & 0.538
  &  0.218 & 0.399 &  0.148 & 0.558
  &  0.165 & 0.247 &  0.108 & 1.181
  &  0.217 & 0.329 &  0.154 & 0.961 \\

MM-PCQA~\cite{zhang2023mmpcqa}
  &  0.155 & 0.184 &  0.105 & 0.540
  &  0.128 & 0.203 &  0.085 & 0.596
  & -0.001 & 0.024 & -0.001 & 1.218
  &  0.122 & 0.190 &  0.081 & 0.999 \\

LMM-PCQA~\cite{zhang2024lmmpcqa}
  & -0.281 & 0.341 & -0.184 & 0.516
  &  0.023 & 0.125 &  0.012 & 0.604
  &  0.085 & 0.159 &  0.056 & 1.203
  &  0.119 & 0.137 &  0.070 & 1.008 \\

CLIP-PCQA~\cite{liu2025clippcqa}
  & -0.114 & 0.160 & -0.075 & 0.542
  &  0.045 & 0.181 &  0.026 & 0.599
  &  0.000 & 0.149 &  0.000 & 1.205
  & -0.021 & 0.019 & -0.017 & 1.017 \\

\bottomrule
\end{tabular}
}

\vspace{2pt}
\begin{minipage}{\textwidth}
\footnotesize
\textit{Note:} SRCC, PLCC, and KRCC denote Spearman rank-order correlation,
Pearson linear correlation, and Kendall rank correlation, respectively;
RMSE denotes root mean squared error. Higher values indicate better
performance for SRCC, PLCC, and KRCC, while lower values indicate better
performance for RMSE.
\end{minipage}

\end{table}